\documentclass[lettersize,journal]{IEEEtran}

\usepackage{amsmath,amssymb,amsopn,amstext,amsfonts}

\usepackage{array}      
\usepackage{tabularx}   
\usepackage{booktabs}
\usepackage{multirow}
\usepackage{multicol}

\usepackage{graphicx}
\usepackage{stfloats}
\usepackage[caption=false,font=normalsize,labelfont=sf,textfont=sf]{subfig}

\usepackage{algpseudocode}  
\usepackage{algorithm}

\usepackage{textcomp}
\usepackage{url}
\usepackage{verbatim}
\usepackage{cite}
\usepackage{pifont}    
\usepackage{utfsym}
\usepackage{bm}

\usepackage[linkcolor=blue,citecolor=blue,urlcolor=blue,colorlinks=true]{hyperref}
\usepackage{cleveref}

\usepackage{tablefootnote}
\usepackage[compatibility=false]{caption}

\graphicspath{{./Figures/}}
\DeclareGraphicsExtensions{.pdf,.png,.jpg,.eps,.svg}
\IEEEoverridecommandlockouts

\title{AeroThrow: An Autonomous Aerial Throwing System for Precise Payload Delivery}
\author{
    Ziliang Li*, Hongming Chen*, Yiyang Lin, Biyu Ye, Ximin Lyu
    \thanks{* means equal contribution.}
    \thanks{Corresponding author: Ximin Lyu}
    \thanks{All authors are with the School of Intelligent Systems Engineering, Sun Yat-sen University, Guangzhou, China. }
    \thanks{E-mail:{\tt\small litleong@mail2.sysu.edu.cn}
    }
}

\begin{document}

\makeatletter
\let\@oldmaketitle\@maketitle
\renewcommand{\@maketitle}{
  \@oldmaketitle
  \begin{center}
    \setcounter{figure}{-1} 
    \includegraphics[width=1.0\linewidth]{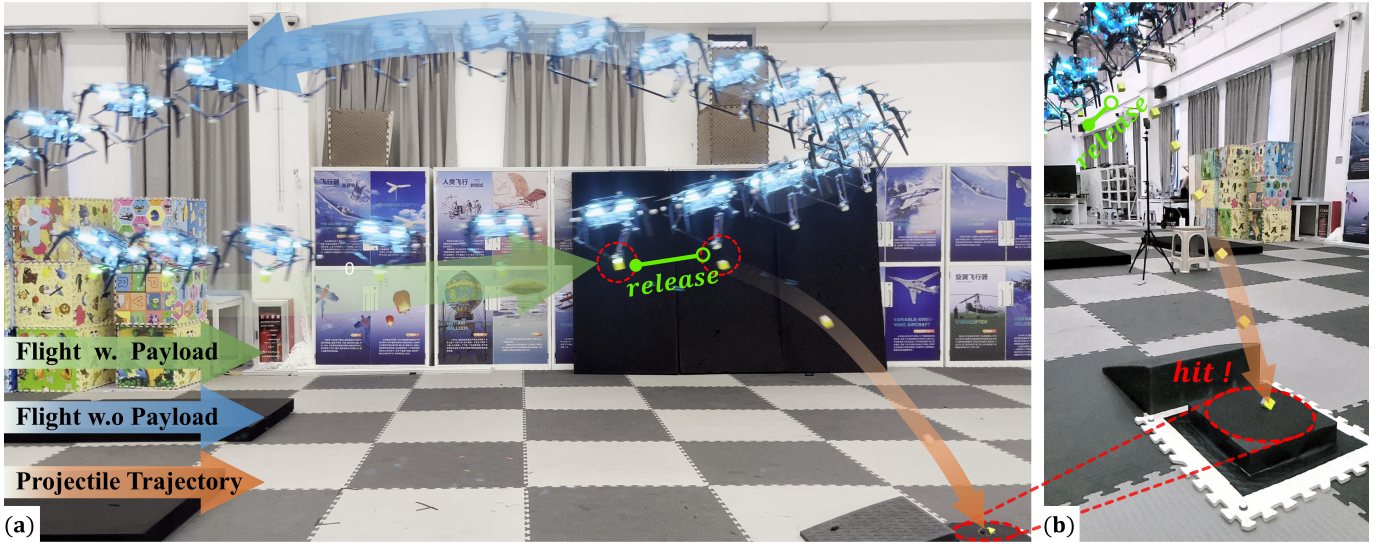}
    \stepcounter{figure}
    
    \vspace{-15pt}
    \captionof{figure}{
      \label{fig:top}
      (a): Agile and precise autonomous aerial throwing. The aerial manipulator carries a payload of unknown mass (200~g in experiments) during flight, releases it with optimal timing, and subsequently returns to its initial position. (b): The payload accurately impacts the designated target location.
    }
    \addcontentsline{lof}{figure}{\protect\numberline{\thefigure}{ 
        (a): Agile and precise autonomous aerial throwing. The aerial manipulator carries a payload of unknown mass (200g in experiments) during flight, releases it with optimal timing, and subsequently returns to its initial position. (b): The payload accurately impacts the designated target location.}
    }
  \end{center}
  \vspace{-20pt}
}
\maketitle
\makeatother

\begin{abstract}
    Autonomous aerial systems play an increasingly vital role in a wide range of applications, particularly for transport and delivery tasks in complex environments. In airdrop missions, these platforms face the dual challenges of abrupt control mode switching and inherent system delays along with control errors. To address these issues, this paper presents an autonomous airdrop system based on an aerial manipulator (AM). The introduction of additional actuated degrees of freedom enables active compensation for UAV tracking errors. By imposing smooth and continuous constraints on the parabolic landing point, the proposed approach generates aerial throwing trajectories that are less sensitive to the timing of payload release. A hierarchical disturbance compensation strategy is incorporated into the Nonlinear Model Predictive Control (NMPC) framework to mitigate the effects of sudden changes in system parameters, while the predictive capabilities of NMPC are further exploited to improve the precision of aerial throwing. Both simulation and real-world experimental results demonstrate that the proposed system achieves greater agility and precision in airdrop missions.

\end{abstract}


\section{Introduction}
    Propelled by the growth of the low-altitude economy and aerial transportation, aerial vehicle systems have been widely used for delivery tasks in challenging scenarios, such as emergency rescue \cite{sanz2022drone} and disaster response \cite{saikin2020wildfire}. In critical support scenarios, airdrop is often essential. However, manual aerial throwing methods, which rely on human operation, are inherently unstable and inefficient. In contrast, autonomous aerial throwing based on UAVs provides a solution with greater flexibility and enhanced accuracy.

    Conventional autonomous aerial throwing systems suffer from poor robustness and limited accuracy. Robustness issues are primarily manifested during airdrop missions, where control mode switching is required during high-speed flight due to payload release, often leading to instability \cite{wu2023ring}. Controllers with fixed parameters can lead to control divergence and pose safety risks. The challenges to the accuracy of aerial throwing are multifaceted. Foremost, precise payload throwing during agile flight maneuvers is highly sensitive to the timing of payload release, posing substantial challenges to actuator control for payload gripping and releasing. Additionally, tracking errors and system delays are intrinsic to UAVs and cannot be systematically eliminated \cite{faessler2017differential}. Even under ideal conditions, the accumulation of such errors can compromise the timing of payload release, thereby exacerbating landing errors \cite{werner2024dynamic}. 
    Current research on aerial throwing does not sufficiently consider the impact of actuator delays and control errors in practical systems on landing accuracy. On one hand, Foehn~et~al.~\cite{foehn2017fast} impose only point-wise parabolic constraints, resulting in high sensitivity to release timing. On the other hand, aerial throwing systems should incorporate adaptive controllers \cite{wu2023ring} or disturbance observers \cite{lyu2018disturbance} to address challenges arising from model variations induced by payload release. Regarding efficiency improvements in airdrop missions, UAV payload systems \cite{tang2015mixed, zeng2020differential, li2023autotrans} offer a viable solution. Although this approach features mechanical simplicity and high payload capacity, the inability to actively and independently actuate the payload compromises system robustness. To address these limitations, we employ an AM with redundant degrees of freedom \cite{cao2024aircrab, chen2025ndob, deng2025whole} for aerial throwing tasks.

    This letter proposes an autonomous aerial throwing system based on an aerial manipulator, where additional actuated degrees of freedom are introduced to compensate for the tracking errors of the UAV. Imposing constraints on the target landing position over a specified time interval during trajectory generation yields both the end-effector trajectory and a feasible release time window, thereby ensuring accurate payload delivery along the reference trajectory. This reference trajectory is then provided as input to an NMPC controller equipped with hierarchical disturbance compensation. Furthermore, leveraging the model predictive property of NMPC, we develop an online method for reassessing the payload release timing. Both simulation and real-world experiments demonstrate that the proposed approach improves landing accuracy across different airdrop trajectories.


    The main contributions of this work can be summarized as follows: 
    \begin{itemize}
    
        \item An autonomous aerial throwing system based on an aerial manipulator is proposed to enable an agile and precise execution of airdrop missions.
        
        \item We propose imposing continuous constraints on the parabolic target landing point, thereby generating a feasible airdrop trajectory that effectively reduces the sensitivity of landing accuracy to payload release uncertainty.
        
        \item By integrating a hierarchical disturbance compensation strategy into the NMPC framework, system robustness and landing accuracy are significantly improved.
    \end{itemize}

\section{Related Work} 
    In recent years, planning and control methods for autonomous airdrop missions have attracted considerable attention, with particular emphasis on trajectory planning, as well as throwing accuracy and precision. Fixed-wing UAVs have been employed in \cite{mathisen2017approach} and \cite{mathisen2020autonomous} to autonomously identify target landing positions and perform trajectory planning and control for airdrop in complex scenarios. However, the maneuverability and throwing accuracy of fixed-wing platforms remain suboptimal. Furthermore, Saikin~et~al.~\cite{saikin2020wildfire} utilizes conventional quadrotors to perform UAV-based firefighting tasks through experimental demonstrations involving object throwing. Several studies have adopted UAV payload systems for aerial throwing and validated their effectiveness via simulation, such as \cite{tang2015mixed} and \cite{zeng2020differential}. Although UAV payload systems exhibit advantages such as high payload capacity and structural simplicity \cite{li2023autotrans}, their cable-suspended configurations prohibit active actuation of the payload, thus compromising system agility and degrees of freedom.

    In UAV airdrop trajectory planning, prior work formulated the combined aerial throwing and UAV planning task as a mixed-integer quadratic programming problem~\cite{faessler2017differential} \cite{foehn2017fast} \cite{tang2015mixed}, and validate their approaches through simulations. Moreover, Foehn~et~al.~\cite{foehn2017fast} and Cao~et~al.~\cite{cao2025time} impose identical landing error constraints within their trajectory optimization processes. However, these constraints limit the reference landing points to a vicinity of the intended target, thereby introducing a first-layer landing error. The second-layer error originates from the controller: releasing the payload abruptly alters the UAV’s model parameters at that instant. Previous studies, such as \cite{panetsos2024nmpc}, employ the Model Predictive Control (MPC) to enhance control performance during airdrop missions. However, uncertain model parameters can lead to inaccuracies in state prediction within MPC, ultimately causing substantial deviations of the actual payload landing position from the planned reference point.

    Previous studies have recently focused on robotic throwing behaviors. Okada~et~al.~\cite{okada2015robust} analyzes the impact of manipulator model uncertainty on throwing precision and optimizes trajectory generation accordingly. In \cite{zeng2020tossingbot}, an end-to-end reinforcement learning framework, named TossingBot, is developed by combining physical simulation and deep learning. It enables robots to learn to grasp and throw arbitrary objects accurately toward a target position directly from visual inputs. However, due to limitations in gripper design, the robot's throwing performance exhibits high sensitivity to the payload release uncertainty. Motivated by \cite{cohen2012state}, works such as \cite{liu2024tube} and \cite{liu2022solution} explicitly exploit throwing task redundancy in robotic trajectory generation, ensuring that payload motion states at the instant of release remain within effective throwing configurations. In addition, Ma~et~al.~\cite{ma2025learning} proposed a learning-based and model-driven framework that achieves accurate whole-body throwing on a legged mobile manipulator. This approach significantly enhances throwing precision and success rates under uncertainty of release.

    Inspired by \cite{liu2024tube}, this letter imposes continuous landing-point constraints over a specified spatio-temporal trajectory interval. By fully exploiting the spatial degrees of freedom inherent in aerial throwing, this method ensures a continuously feasible release-time interval in the trajectory output, thus reducing sensitivity of throwing performance to payload release uncertainty. In terms of control design, we integrate hierarchical disturbance compensation into the NMPC framework, significantly enhancing control performance when carrying payloads of unknown mass.

\section{Preliminaries} 
    \subsection{Notation}
        The aerial throwing platform utilized in this letter consists of a quadrotor and a delta arm. Given the negligible distance between the end-effector and the payload, the state of the payload immediately prior to release is approximated by that of the end-effector. To accurately describe the proposed system, three right-handed coordinate frames are introduced: 
        \begin{enumerate}[]
            \item \textit{Inertial Frame} $\mathcal{\bm{F}}_I:\{x_I,y_I,z_I\}$ with $z_I$ pointing upward, opposite to gravity; 
            
            \item \textit{Body Frame} $\mathcal{\bm{F}}_B:\{x_B,y_B,z_B\}$ with $x_B$ pointing forward and $z_B$ aligned with the collective thrust direction; 
            
            \item \textit{End-Effector Frame} $\mathcal{\bm{F}}_E:\{x_E,y_E,z_E\}$ with its origin at the end-effector and related to the Body Frame by translation only. 
        \end{enumerate}

\begin{figure}[t]
    \vspace{-0.3cm}
        \centering
            \includegraphics[width=0.9\linewidth]{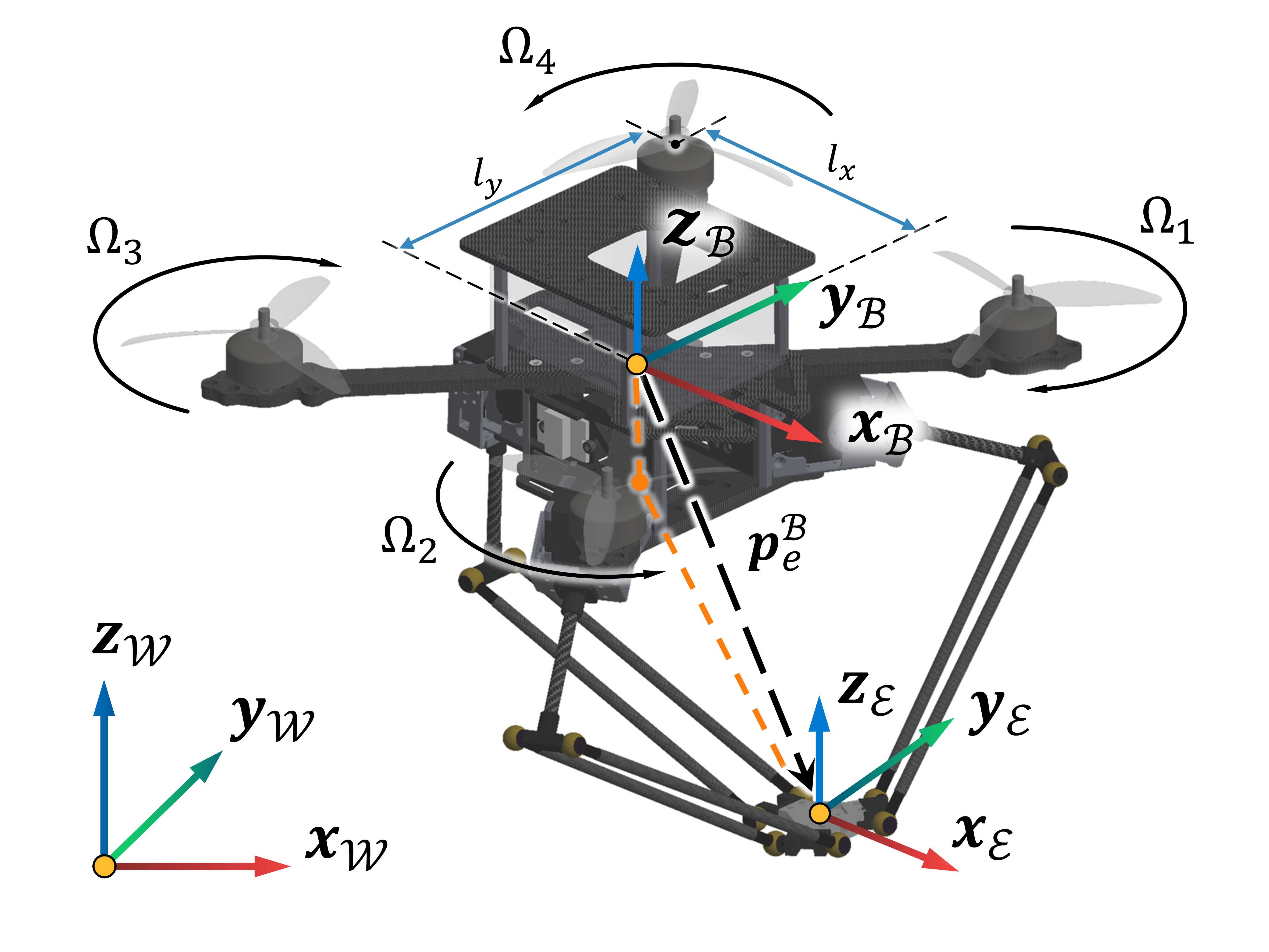}
            \caption{Coordinate definitions, relevant physical parameters, and notations.}
        \label{fig:frame}
    \vspace{-0.3cm}
    \end{figure}

        The rotation from $\mathcal{F}_I$ to $\mathcal{F}_B$ is represented by the rotational matrix $\bm{R}(\bm{q}) = [x^B,\, y^B,\, z^B] \in \mathrm{SO}(3)$ parameterized by quaternion $\bm{q} = [q_w,\, q_x,\, q_y,\, q_z] ^ T \in \mathbb{S}^3$.  Throughout this letter, the subscripts indicate the physical meaning of each variable, while the superscript denotes the coordinate frame in which the vector is expressed. The relevant physical parameters and notations are illustrated in Fig.\ref{fig:frame}.

    \subsection{Aerial Manipulator Model}
            In this work, the state and input of the quadrotor platform are described using the flat outputs at the end-effector center and their derivatives. Based on the structural properties of the delta arm, the rotational dynamics at the end-effector center are approximated to be consistent with those of the quadrotor platform. The pendulum effect induced by rotation in $\mathcal{\bm{F}}_E$ requires a change in the translational state of the quadrotor platform. The relationship between these quantities is established as follows:
            {\small
            \begin{subequations}
                \begin{align}
                    \bm{p}_{q}^{W} = \ \bm{p}_{e}^{W} & - \bm{R}(\bm{q}) \bm{p}_{e}^{B}, \label{eq:flatness_a} \\ 
                    \dot{\bm{p}}_{q}^{W} = \ \dot{\bm{p}}_{e}^{W} & - \bm{R}(\bm{q}) \dot{\bm{p}}_{e}^{B} - \widehat{\bm{\omega}} \Big(\bm{R}(\bm{q}) \bm{p}_{e}^{B}\Big), \label{eq:flatness_b} \\ 
                    \ddot{\bm{p}}_{q}^{W} = \ \ddot{\bm{p}}_{e}^{W} & - \bm{R}(\bm{q}) \ddot{\bm{p}}_{e}^{B} 
                    - 2\widehat{\bm{\omega}} \Big(\bm{R}(\bm{q}) \dot{\bm{p}}_{e}^{B}\Big) \notag \\
                    & - \widehat{\bm{\omega}} \Big(\bm{R}(\bm{q}) \bm{p}_{e}^{B}\Big) - \widehat{\bm{\omega}}^2 \Big(\bm{R}(\bm{q}) \bm{p}_{e}^{B}\Big), \label{eq:flatness_c}
                \end{align}
                \label{eq:flatness}
            \end{subequations}
            }\unskip
            where $\bm{p}_{q}^{W}$ and $\bm{p}_{e}^{W}$ are the center position of the quadrotor and end-effector.  $\widehat{\bm{\omega}}$ represents the skew-symmetric matrix associated with $\bm{\omega}^B$, which denotes the angular velocity of the AMs in $\mathcal{F}_B$. 
        
            Similar to \cite{faessler2017differential}, \cite{sun2022comparative}, the quadrotor translational dynamic is formulated as 
            \begin{equation}
                \begin{split}
                    \ddot{\bm{p}}_{q}^{W} &= (T \bm{z}_{B} + \bm{f}^{W}_\text{ext}) / m + \bm{g},
                \end{split}
                \label{eq:translational}
            \end{equation}
            where $T$ and $m$ represent the collective thrust and the AMs mass; $\bm{g} \in \mathbb{R}^{3}$ is the gravitational vector; $\bm{z}_{B}$ is the z-axis of the body frame in the inertial frame; and $\bm{f}^{W}_\text{ext}$ indicates the force caused by external disturbance.

            The rotational kinematic and dynamic equations are expressed as 
            {\small
            \begin{subequations}
                \begin{align}
                    \dot{\bm{q}} &= \frac{1}{2}\bm{q} \otimes \begin{bmatrix}
                                                                0 \\
                                                                \bm{\omega}^B
                                                              \end{bmatrix}\label{eq:rot_1}, \\
                    \dot{\bm{\omega}}^B &= \mathcal{I}^{-1} (\bm{\tau} - \bm{\omega}^B \times \mathcal{I} \bm{\omega}^B + \bm{\tau}^B_\text{ext})\label{eq:rot_2},
                \end{align}
                \label{eq:rotational}
            \end{subequations}
            }\unskip
            where $\otimes$ denotes the quaternion multiplication operator. $\bm{\tau}$ and $\mathcal{I}$ represent the resultant torque generated by rotors and the inertia matrix. The model uncertainties and external disturbance in the body torque are denoted by $\bm{\tau}^B_\text{ext}$.
        
    \subsection{Projectile Motion Model}
        In this section, the projectile motion of the payload after release is modeled. The landing point $\bm{P}_l$ is determined by the position $\bm{p}_l^{W}(t_\text{r})$ and its derivative of the payload at the instant of release, while the release time is represented by $t_\text{r}$. The time $T_r$ for the payload to reach the horizontal plane of the target point $\bm{p}^W_t$ is calculated as follows: 
        {\small
        \begin{equation}
            \begin{split}
                T_r = \frac{\dot{z_l}(t_\text{r}) + \sqrt{\dot{z_l}(t_\text{r})^2 + 2gz_l(t_\text{r})}}{g}, 
            \end{split}
            \label{eq:drop_time}
        \end{equation}
        }where $z_l(t_\text{r})$ and $\dot{z_l}(t_\text{r})$ are the release height and vertical velocity of the payload; $g$ denotes the absolute value of gravitational acceleration. The trajectory of the payload as a function of time $t$ is formulated as: 
        {\small
        \begin{equation}
            \begin{split}
                \bm{p}_l^{W}(t) = 
                    \begin{cases}
                        \bm{p}_e^{W}(t), & \text{if } 0 \leq t \leq t_\text{r} \\
                        \bm{p}_l^{W}(t_\text{r}) + \dot{\bm{p}}^{W}_lt + \frac{1}{2}\bm{g}t^2, & \text{if } t > t_\text{r}
                    \end{cases} 
            \end{split}
            \label{eq:drop_traj}
        \end{equation}
        }

        Consequently, the landing point $\bm{P}_l$ can be expressed as: 
        \begin{equation}
            \begin{split}
                \bm{P}_l &= \bm{p}_l^{W}(t_r + T_r). 
            \end{split}
            \label{eq:land_point}
        \end{equation}

        In the foregoing analysis, a simplified projectile motion model is adopted. Specifically, it is assumed that the payload is subject to no external forces in the horizontal direction and thus maintains a constant velocity equal to its value at the moment of release. Meanwhile, the vertical motion is governed exclusively by gravitational acceleration. 
    
    \section{Methodologies}
\begin{figure*}[t]
            \vspace{-20pt}
            \centering
            \includegraphics[width=1\linewidth]{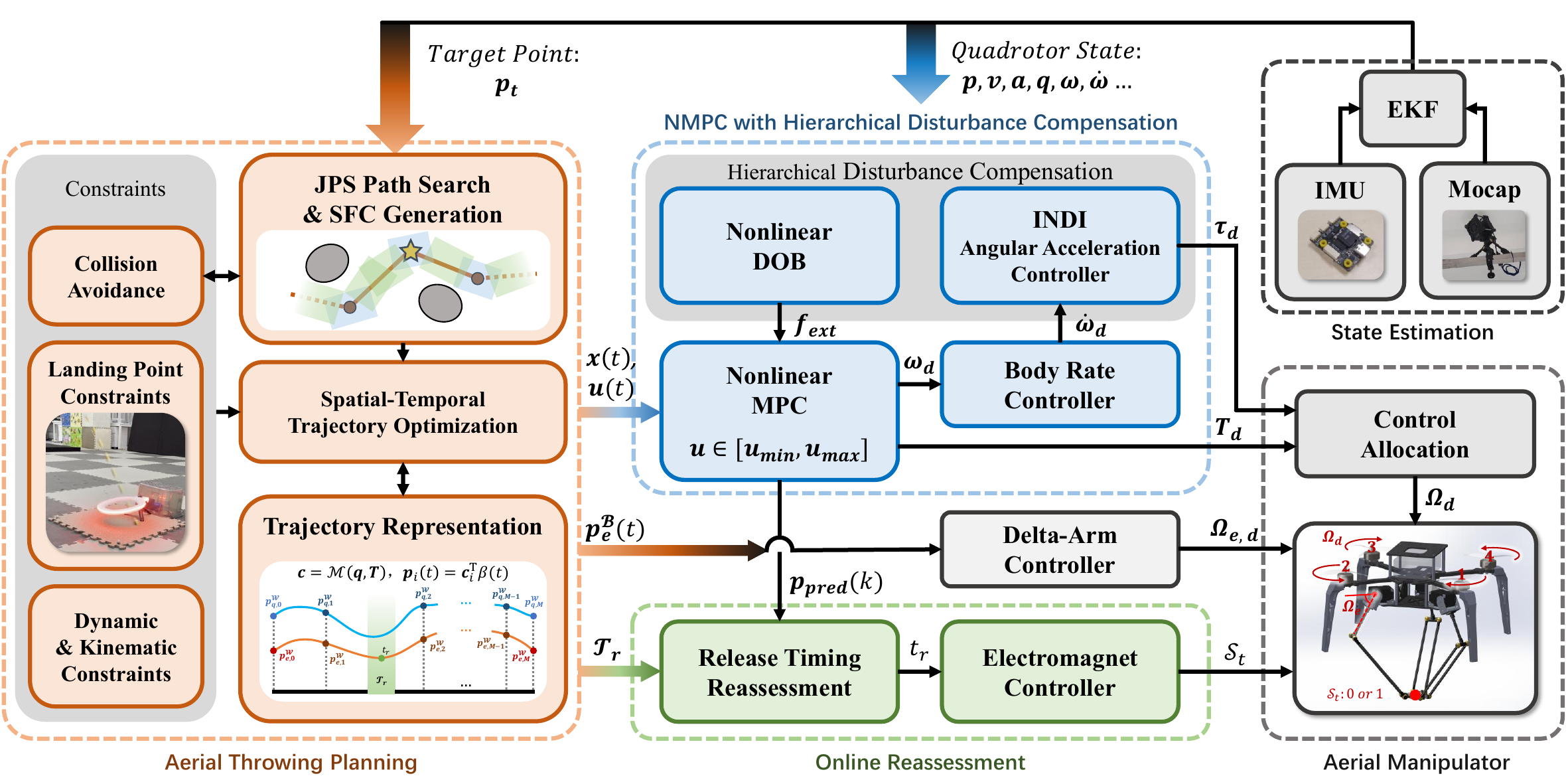}
            \caption{Overview of our autonomous aerial throwing system. (a) The Aerial Throwing Planning module generates reference trajectories for both the UAV and the delta arm. (b) The NMPC framework incorporates a hierarchical disturbance compensation strategy. (c) Based on model state predictions, the Online Reassessment module utilizes the reference release time window to update the payload release timing online and control the electromagnet accordingly.}
            \label{fig:pipeline}
            \vspace{-0.2cm}
        \end{figure*}

    The overall architecture of the proposed autonomous airdrop system is depicted in Fig.~\ref{fig:pipeline}. We identify three primary sources of error in airdrop missions: reference landing error from the planner, tracking error from the controller, and system delays. The following subsections describe our approaches to addressing these issues at the levels of trajectory planning (Sec.~\ref{sub:Plan}), control framework design (Sec.~\ref{sub:ctrl}), and payload release trigger logic (Sec.~\ref{sub:reass}), respectively. 
    
        \subsection{Aerial Throwing Trajectory Planning}\label{sub:Plan}
        \subsubsection{Nonlinear Optimization Construction}
        $M$ pieces of $D$\-dimension polynomial spline $\bm{p}(t) = \{\bm{p}_1(t), ..., \bm{p}_M(t)\}$ with degree $N = 2s - 1$ are used to represent the whole flat-output trajectories and $s$ are the order of the relevant integrator chain. We utilize $\bm{T} = \{T_1, ..., T_M\} \in \mathbb{R}^M_{>0}$ represents the durations of all pieces and $T_{\sigma} = \sum_{i = 1}^{M} T_i$ denotes the sum of time. For the $i\text{-th}$ segment of the trajectory, $\bm{p}_i(t)$ is defined as: 
            \begin{equation}
                \begin{split}
                    \bm{p}_i(t) = \bm{c}_i^T \bm{\beta}(t), 
                \label{eq:pi(t)}
                \end{split}
            \end{equation}
        where $i \in \{1, 2, ..., M\}$, $\bm{c}_i \in \mathbb{R}^{2s \times D}$ is the coefficient matrix for each piece, and $\bm{\beta}(t) = [t^0, t^1, ..., t^D]$ is the natural basis. 

        The MINCO framework~\cite{wang2022geometrically} is widely used in trajectory optimization due to its efficiency in generating smooth trajectories with minimal control efforts. We adopt this method and formulate a continuous and differentiable penalty function to enforce landing-point constraints, ensuring trajectory feasibility and smoothness. MINCO maps the original polynomial coefficients $\bm{c}$ to intermediate waypoints $\bm{q}$ and time allocations using the parameter mapping constructed $\bm{c} = \mathcal{M} (\bm{q}, \bm{T})$, where $\bm{q} =~\{q_1, ..., q_{M-1}\}$, and $\mathcal{M}$ is a smooth and linear-complexity map. 

        The trajectory optimization problem can be formulated as follows:
        {\small
        \begin{subequations}
            \begin{align}
                \operatorname*{min}_{\bm{p}(t), ~\bm{T}} &\mathcal{J} = \int_0^{T_\sigma} \|\bm{p}^{(s)}(t)\|^2 + \rho T_\sigma \label{eq:minco_func} \\
                ~\text{s.t.} \quad 
                &\mathcal{G}(\bm{p}(t), ..., \bm{p}^{(s)}(t)) \preceq \bm{0}, ~\forall t \in [0, T_\sigma] \label{eq:penalty} \\
                &\bm{p}^{[s-1]}(0) = \bm{p}_o, ~\bm{p}^{[s-1]}(T_M) = \bm{p}_f \label{eq:init_term} \\
                &T_i > 0, \ i \in \{1, 2, ..., M\}, \label{eq:T_i}
            \end{align}
            \label{eq:MINCO}
        \end{subequations}
        }\unskip
        where Eq.~\eqref{eq:minco_func} minimizes the control energy involving a time regularization. Equations~\eqref{eq:penalty} and \eqref{eq:init_term} represent inequality and equality constraints, encompassing corridor, kinematic, dynamic, and landing-point constraints. $\mathcal{G}$ is the inequality constraints over $[0, T_\sigma]$. $\bm{p}_o$ and $\bm{p}_f$ denote the initial and terminal conditions for the quadrotor and the end-effector.

        \subsubsection{Constraints Formulation}
            We follow the methodology presented in \cite{deng2025whole} to model the corridor, kinematic, and dynamic constraints in a unified manner. Inspired by \cite{ji2022real}, the landing error of the payload can be constrained by constructing such a penalty function: 
            \begin{equation}
                \begin{split}
                    \mathcal{G}_l(t) = \mathcal{L}_\mu[\mathcal{E}_t  + \tau] \|\bm{P}_l - \bm{p}^W_t\|^2, 
                \end{split}
                \label{eq:landing_penalty}
            \end{equation}
            where $\mathcal{E}_t = 1 - |t - t_r|$ and $\tau$ denotes the time interval, $\mathcal{L}_\mu$ is a smoothing of the landing penalty which is indicated by: 
            {\small
                \begin{equation}
                    \begin{split}
                        \mathcal{L}_\mu [x] = 
                        \begin{cases}
                            0, &x \leq 1 -2\mu\\
                            \frac{1}{2 \mu^4} (x + 2\mu - 1)^3 (1 - x), &1 -2\mu < x \leq 1 - \mu\\
                            \frac{1}{2 \mu^4} (x - 1)^3 (x + 2\mu - 1) + 1, &1 - \mu < x \leq 1\\
                            1, &x > 1
                        \end{cases}
                    \end{split}
                    \label{eq:smooth}
                \end{equation}
            }where $\mu$ controls the smoothness of the relaxation function. $\mathcal{L}_\mu$ smoothly incorporates integer variables into the nonlinear programming model through a defined relaxation function. This formulation does not constrain the position of release, but rather lets the optimization process find the best value. 

\begin{figure}[t]
                \centering
                \vspace{-15pt}
                \includegraphics[width=0.9\linewidth]{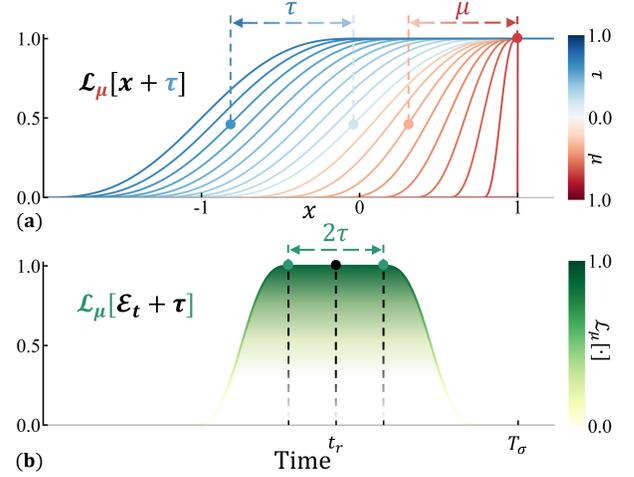}
                \caption{Visualization of the relaxation function. (a): The transition of the activation strength from 0 to 1 as $\mu$ and $\tau$ vary is depicted. (b): The time-varying activation of the penalty term is presented, where the objective is that all flat states generated within the time interval of maximum activation strength, $\mathcal{\bm{T}}_r$, satisfy the payload release condition.}
                \label{fig:drop_smooth}
            \end{figure}

            Considering the actuator response delay in controlling the payload gripper, it is insufficient to impose landing constraints at a single release instant during optimization. To reduce the sensitivity of throwing accuracy to the precise timing of the payload release, we apply identical landing penalties throughout the time interval $\tau$ surrounding the optimal release time $t_r$ identified during the optimization process. As shown in Fig.~\ref{fig:drop_smooth}, this enables the planner to fully exploit the spatial redundancy of the throwing task and generate a trajectory that contains a continuous set of feasible release instants $\mathcal{\bm{T}}_r = \{t|t \in [t_r - \tau, t_r + \tau]\}$. 

    \subsection{Controller with Hierarchical Disturbance Compensation}\label{sub:ctrl}
    
        \subsubsection{Nonlinear Model Predictive Controller (NMPC)}
            The NMPC generates control commands by solving a finite-time optimal control problem in a receding horizon framework. Given the reference throwing trajectory, the cost function is formulated based on the deviation between the predicted and reference states over the time horizon, thereby incorporating multiple reference points within the horizon.
            
            We consider the state of the quadrotor as $\bm{x} = [\bm{p}^W, \bm{v}^W, \bm{q}]^T$ and the control input as $\bm{u} = [T, \bm{\omega}^B]^T$. The time horizon $\tau_h \in~[t, t + h]$ is divided into $N$ equal intervals of size $dt = h/N$, with $h$ denoting the horizon length. This discretization yields a constrained nonlinear optimization problem:
            {\small
                \begin{equation}
                    \begin{split}
                        \bm{u}_\text{NMPC} = \operatorname*{\arg\min}_{\bm{u}} 
                            \sum_{k = 0}^{N - 1} \big( \| \bm{x}(k) - \bm{x}_r(k) \|_{\bm{Q}} 
                            + \| \bm{u}(k) - \bm{u}_r(k) \|_{\bm{Q}_u} \big) \\
                            + \| \bm{x} (N) - \bm{x}_r (N) \|_{\bm{Q}_N} \\
                        \text{s.t.} \quad 
                        \bm{x}(k-1) = \bm{f}(\bm{x}(k), \bm{u}(k)), \ 
                        \bm{x}_0 = \bm{x}_\text{now},\ 
                        \bm{u} \in [\bm{u}_\text{min}, \bm{u}_\text{max}].
                    \end{split}
                    \label{eq:ocp}
                \end{equation}
            }

            The function $\bm{f}(\bm{x}(k), \bm{u}(k))$ is the quadrotor model given by Equations~\eqref{eq:translational} and \eqref{eq:rotational}. The reference state $\bm{x}_r$ and input $\bm{u}_r$ are derived from trajectories. $\bm{Q} = diag(\bm{Q}_p, \bm{Q}_v, \bm{Q}_q)$, $\bm{Q}_u$ and $\bm{Q}_N$ are the state, input and terminal state weighting matrices. $\bm{x}_\text{now}$ is the current state estimation when solving the optimal control problem. The ACADO \cite{verschueren2018towards} toolkit, together with qpOASES \cite{ferreau2014qpoases}, is used as the solver for the nonlinear optimization algorithm. This nonlinear quadratic optimization problem can be solved using a real-time iteration scheme. 
        
        \subsubsection{Hierarchical Disturbance Compensation}
            Given the variability and uncertainty of the payload mass during airdrop missions, a nonlinear disturbance observer (NDOB) is implemented to compensate for external disturbance forces affecting the system. The ability of this approach to improve the robustness of the system has been substantiated in previous studies~\cite{chen2025ndob}~\cite{yu2024dob}. 

            We can obtain: 
            \begin{equation}
                \begin{split}
                    \bm{f}^W_\text{ext} = m(\bm{a}^W - \bm{g}) - T\bm{z}_B,
                \end{split}
                \label{eq:ndob_dyn}
            \end{equation}
            where $\bm{a}^W$ represents the acceleration derived from the IMU. Subsequently, the observer matrix is constructed \cite{yu2024dob}, followed by differentiation and discretization, yielding:
            {\small
            \begin{equation}
                \begin{split}
                    \bm{f}^W_\text{ext}(\tau + 1) = \bm{f}^W_\text{ext}(\tau) + \frac{c}{m} (m\bm{a} - m\bm{g} - T\bm{z}_b - \bm{f}^W_\text{ext}) \ {\Delta t}.
                \end{split}
                \label{eq:ndob_ob}
            \end{equation}
            }

            Here, $\bm{f}^W_\text{ext}$ is filtered with a Butterworth filter at the cut-off frequency (50Hz). $\Delta t$ denotes the reciprocal of the sensor frequency. In order to achieve rapid convergence of the disturbance force estimate, the parameter $c$ should be adjusted based on the actual operating conditions. The thrust $T$ and torque vector $\bm{\tau}$ can be obtained through the angular velocity $\bm{\Omega}$ of each rotor by the model of quadrotor, as shown in Fig.~\ref{fig:frame}: 
            {\small\begin{subequations}
            \begin{align}
                \begin{bmatrix} T \\ \bm{\tau} \end{bmatrix} &= \bm{G}_1 {\bm{\Omega}^\circ}^2 + \bm{G}_2 \dot{\Omega}\label{eq:T_tau}, \\
                \bm{G}_1 &= \begin{bmatrix} c_t & c_t & c_t & c_t \\ 
                                            l_yc_t & -l_yc_t & -l_yc_t & l_yc_t \\ 
                                            -l_xc_t & -l_xc_t & l_xc_t & l_xc_t \\ 
                                            -c_m & c_m & -c_m & c_m  \end{bmatrix},\label{eq:G1} \\ 
                \bm{G}_2 &= \begin{bmatrix} 0 & 0 & 0 & 0 \\
                                            0 & 0 & 0 & 0 \\ 
                                            0 & 0 & 0 & 0 \\ 
                                            \mathcal{I}_r & -\mathcal{I}_r & \mathcal{I}_r & -\mathcal{I}_r \end{bmatrix}\label{eq:G2},
            \end{align}
            \label{eq:quad_mat}
            \end{subequations}}\unskip
            where $l$ is the size of the quadrotor, $c_t$ and $c_m$ denote the thrust coefficient and the torque coefficient, respectively. $\mathcal{I}_r$ is the inertia of the propeller and $^\circ$ indicates the Hadamard power. 

            Prior to release, the payload remains affixed at a location offset from the quadrotor's center of mass, thereby continuously generating an additional moment acting on the system. As the moment arm depends on the manipulator's configuration and is challenging to measure both in real time and precisely, we draw inspiration from \cite{tal2020accurate} to implement an incremental nonlinear dynamic inversion (INDI) scheme within the inner-loop angular velocity controller, enabling effective compensation for this external moment. 

            The desired angular acceleration $\dot{\bm{\omega}}^B_r$ is calculated as 
            {\small
            \begin{equation}
                \begin{split}
                    \dot{\bm{\omega}}^B_d = \mathcal{K} (\bm{\omega}^B_r - \bm{\omega}^B_f) + \dot{\bm{\omega}}^B_r, 
                \end{split}
                \label{eq:pd_omega}
            \end{equation}
            }where $\bm{\omega}^B_r$ and $\dot{\bm{\omega}}^B_r$ denote the reference angular velocity and acceleration, $\bm{\omega}^B_f$ is the filtered feedback angular velocity from IMU. Then, from INDI control law \cite{tal2020accurate}, we have 
            {\small
            \begin{equation}
                \begin{split}
                    \bm{\tau}_d = \bm{\tau}_f + \mathcal{I} (\dot{\bm{\omega}}^B_d - \dot{\bm{\omega}}^B_f), 
                \end{split}
                \label{eq:indi}
            \end{equation}
            }where $\dot{\bm{\omega}}^B_f$ denotes the filtered feedback angular acceleration, which is obtained as the derivative of $\bm{\omega}^B_f$. The $\bm{\tau}_f$ is the feedback angular moment from the angular velocity of each propeller based on Eq.~\eqref{eq:G2}. Finally, from Eq.~\eqref{eq:T_tau}, we can solve the equation with the control thrust and the control moment commands: 
            {\small
            \begin{equation}
                \begin{split}
                    \begin{bmatrix}
                        T_d \\ \bm{\tau}_d
                    \end{bmatrix} = \bm{G}_1 {\bm{\Omega}^\circ_d}^2 + {\Delta t}^{-1} \bm{G}_2 (\bm{\Omega}_r - \bm{\Omega}_f), 
                \end{split}
                \label{eq:ctrl_allocate}
            \end{equation}
            }where $\bm{\Omega}_f$ is the current rotor angular velocity and $\Delta t$ is the motor dynamics time constant. To mitigate the influence of IMU noise on sensor measurements, both $\bm{\omega}^B_f$ and $\bm{\Omega}_f$ are processed through second-order Butterworth filters with a same cut-off frequency. 

    \subsection{Online Release Timing Reassessment}\label{sub:reass}
        Previous studies on autonomous airdrop typically initiate payload release at the nominal time specified by a high-level planner or by updating the reference release timing at relatively low frequencies. However, accumulation of system delays and control errors can compromise the optimality of this reference time. In this subsection, we propose a release timing reassessment strategy that leverage the predictive capabilities of NMPC.

        Although our method yields a short feasible interval for releasing, it still requires a precise triggering time for payload release. We posit that the reference release timing $t_r$ along the reference airdrop trajectory $p_\text{ref}(t)$ yields the optimal landing location. However, in the presence of various disturbances, the actual optimal release time $t_r^{*}$ typically lies within a neighborhood of $t_r$. It is assumed that: 
        \begin{equation}
            \begin{split}
                |t_r^{*} - t_r| \leq h. 
            \end{split}
            \label{eq:assum}
        \end{equation}
        
        Accordingly, from the moment the reference release time enters the NMPC's prediction horizon until the current time coincides with the reference release time, the prediction window consistently contains the optimal release point along the current flight trajectory $p(t_r^{*})$.

        To fully exploit the model predictive property, we assume that the payload is released at each discrete state $p_{pred}(k)$ within the prediction horizon. According to Equations~\eqref{eq:drop_time} and \eqref{eq:land_point}, this yields a set of predicted landing locations corresponding to the current time: 
        {\small
        \begin{subequations}
            \begin{align}
                \bm{T}_{r,~i} &= \frac{\dot{z}_{pred}(i) + \sqrt{\dot{z}_{pred}(i)^2 + 2gz_{pred}(i)}}{g}\label{eq:pred_t}~(1 \leq i \leq N), \\ 
                \bm{P}_l(i) &= \bm{p}_{l,~i}^W\big(t_i + T_r(i)\big)\label{eq:pred_p} ~(1 \leq i \leq N).
            \end{align}
            \label{eq:landing_list}
        \end{subequations}
        \vspace{-0.5cm}
        }

        This allows the corresponding landing error sequence, denoted as $\mathcal{E}$, to be calculated: 
        \begin{equation} 
            \begin{split} 
                \mathcal{E}(i) = \| \bm{P}_l(i) - \bm{p}^W_t \|~(1 \leq i \leq N).
            \end{split}
            \label{eq:error_list}
        \end{equation}
        Since the planned reference landing point may not exactly coincide with the desired target location, we bypass the reference landing point and directly compute the error between the predicted landing point and the target location.

        Based on the predicted sequence of landing errors, incorrect release logic may result in missing or misidentifying the optimal release timing. To address this, we design an efficient decision mechanism that dynamically searches for the optimal release instant in real time, as detailed in Alg.~\ref{Algorithm:1}. The method continuously updates $t_r$ to minimize landing errors during flight. By appropriately setting the stopping condition, it also accounts for actuator delays and prevents Zeno behavior.
        
        \begin{algorithm}
        \caption{Release Timing Reassessment}
            \begin{algorithmic}[1] 
            
                \While{$t_r - h \leq \tau \leq t_r + h$}
                    \State \text{Compute}~$\mathcal{E}$~\text{from the current predicted state sequence}
                    \State $k^* \leftarrow \arg\min_{k} \mathcal{E}[k]$
                    \State $\Delta t \leftarrow k^* \cdot dt$
                    \State $t_r \leftarrow t_{\text{now}} + \Delta t$
                    \If{$\Delta t \leq dt$}
                        \State \textbf{break}
                    \EndIf
                \EndWhile
                
            \end{algorithmic}
            \label{Algorithm:1}
        \end{algorithm}
        
\section{Experiments}
    In this section, we evaluate the performance of our autonomous aerial throwing system through simulation and real-world experiments. The experimental validation comprises three parts: 1) trajectory generation validation for aerial throwing with continuous landing-point constraints demonstrated reduced sensitivity to the release uncertainty (Sec.~\ref{sub:exp1}), 2) ablation studies of the hierarchical disturbance compensation framework, which illustrate the robustness of the control system (Sec.~\ref{sub:exp2}), and 3) comparative experiments on system throwing precision conducted under various airdrop flight trajectories, validating the effectiveness of the proposed method in mitigating inherent errors such as system delays (Sec.~\ref{sub:exp3}).

\begin{figure}[h]
        \centering
        \includegraphics[width=1.0\linewidth]{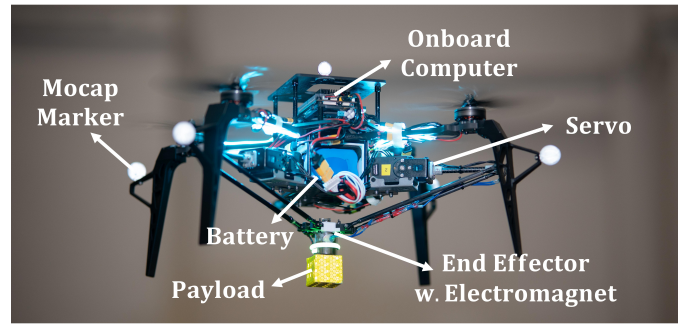}
        \caption{Real-world aerial manipulator platform used for experiment.}
        \label{fig:exp_conf}
        \vspace{-10pt}
    \end{figure}

    Our system is developed in C++11 on Ubuntu 20.04 with ROS Noetic. The experimental platform, depicted in Fig.\ref{fig:exp_conf}, features a quadrotor equipped with a delta arm, an NVIDIA Jetson Orin NX 16GB as the onboard computer, and an NxtPX4v2 flight controller. With the addition of a 6S battery, the total mass of the platform is approximately 1.59kg. The control design for the delta arm follows the methodology described in \cite{chen2025ndob}. The delta arm is actuated by three DYNAMIXEL XL430-W250-T servo motors and fitted with an electromagnet at its end-effector. The electromagnet is engaged to hold the payload during flight, and its magnetic field is immediately deactivated to release the payload when the current time reaches $t_r$. For state estimation, an Extended Kalman Filter (EKF) fuses measurements from the NOKOV Motion Capture System and the onboard IMU to provide robust and accurate system states.
    \subsection{Planning Performance}\label{sub:exp1}
\begin{figure}[t]
        \centering
        \includegraphics[width=1.0\linewidth]{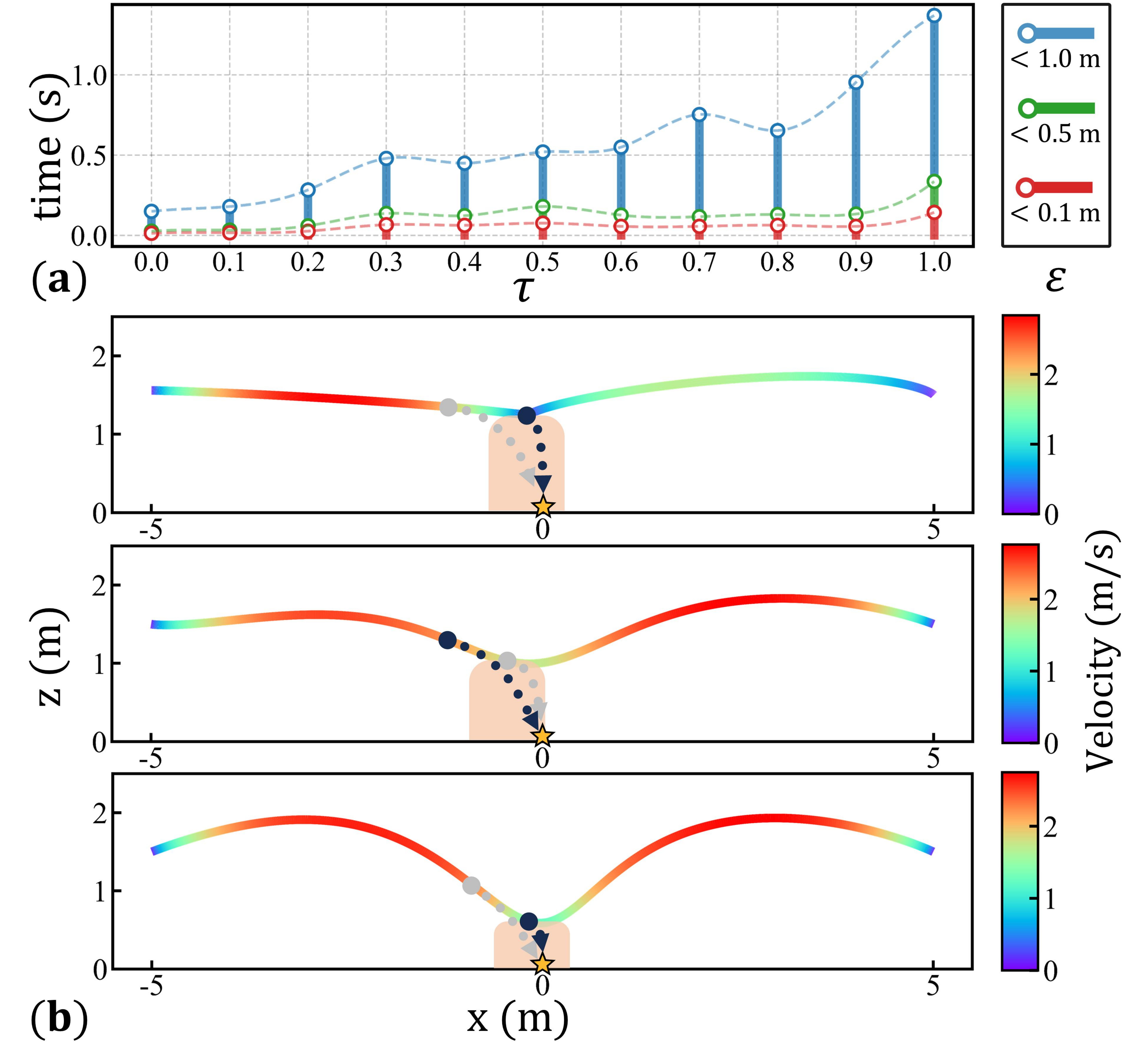}
        \caption{(a): The solid lines of different colors represent the durations for which the landing error along the trajectory remains below different threshold values, while the dashed lines show how these durations vary with the parameter $\tau$. (b): By varying the smoothing parameter $\mu$, it can be observed that the action of releasing the payload gradually transitions to nearly \textit{place}.}
        \label{fig:exp_1}
        \vspace{-10pt}
    \end{figure}

        In this section, we evaluate the performance of the planning module in the executing airdrop missions, particularly emphasizing the trajectory’s sensitivity to release uncertainty. By incrementally increasing the parameter $\tau$ in the relaxation function~\eqref{eq:smooth} during trajectory optimization, we observe that the feasible throwing interval within the optimized airdrop trajectory is correspondingly extended. This effect primarily arises because the planner reduces the flight speed during the time interval in which the penalty is activated, eventually causing the UAV to nearly hover directly above the target location. Consequently, the task execution shifts from a $\textit{}{throw}$ action, primarily driven by horizontal velocity, to a $\textit{drop}$ action, resulting in reduced sensitivity of the airdrop trajectory to release uncertainty, as shown in Fig.~\ref{fig:exp_1}~(a). However, the duration of full activation of the relaxation function must not exceed a certain proportion of the total flight time; otherwise, it would severely compromise the smoothness of the trajectory states.

        By adjusting the value of $\mu$, the optimizer modifies the flight state to facilitate the airdrop, as shown in Fig.~\ref{fig:exp_1}~(b). Lowering the flight altitude shortens the free-fall duration, enabling more precise landing with minimal change to flight velocity. Ultimately, this approach exhibits a trend in the release action shifting from a $\textit{throw}$ toward a $\textit{place}$, each representing a distinct mode of payload release.



    \subsection{Controller Ablation Study}\label{sub:exp2}
        In this section, we evaluate the performance of the proposed NMPC control framework through ablation studies of its modules in three scenarios with different parabolic velocities. To assess the robustness of the controller in these three airdrop flight scenarios, we introduce a payload with an unknown mass of 200g and release it strictly at the reference release timing. Landing point coordinates are collected in real-world flight, and the feasibility of the proposed NMPC framework in improving the concentration of landing points is measured by the dispersion of the actual landing points compared to the reference. Performance is quantified using RMSE/MAX, as presented in Tab.~\ref{tab:exp_2}, where $v_r$ denotes the reference release velocity specified by the trajectory.

        \begin{table}[b]
            \centering
            \caption{NMPC Ablation Study in Real Environment}
            \renewcommand{\arraystretch}{1.1}
            \setlength{\tabcolsep}{4pt}  
            \begin{tabular}{cc|cccc}
            \toprule
            \multicolumn{2}{c|}{Trajectory}
            & \multicolumn{4}{c}{Landing Error (RMSE/MAX) [cm]} \\
            \midrule
            \begin{tabular}{c} $v_{r}$ \\ {[m/s]} \end{tabular}
            & \begin{tabular}{c} $a_{\max}$ \\ {[m/s$^2$]} \end{tabular}
            & NMPC & w. NDOB & w. INDI & \begin{tabular}{c} w. NDOB \\ \& INDI \end{tabular} \\
            \midrule
            4.9 & 5.0 & 15.4/20.2 & 6.7/10.5 & 13.6/21.2 & \textbf{5.9/8.9} \\
            2.2 & 3.0 & 22.0/48.3 & 9.7/12.7 & 13.6/17.1 & \textbf{5.2/7.2} \\
            2.5 & 2.8 & 4.1/6.4   & 1.9/2.0  & 2.9/3.7   & \textbf{1.5/2.0} \\
            \bottomrule
            \end{tabular}
            \label{tab:exp_2}
        \end{table}
        
\begin{figure}[h]
            \centering
            \includegraphics[width=1.0\linewidth]{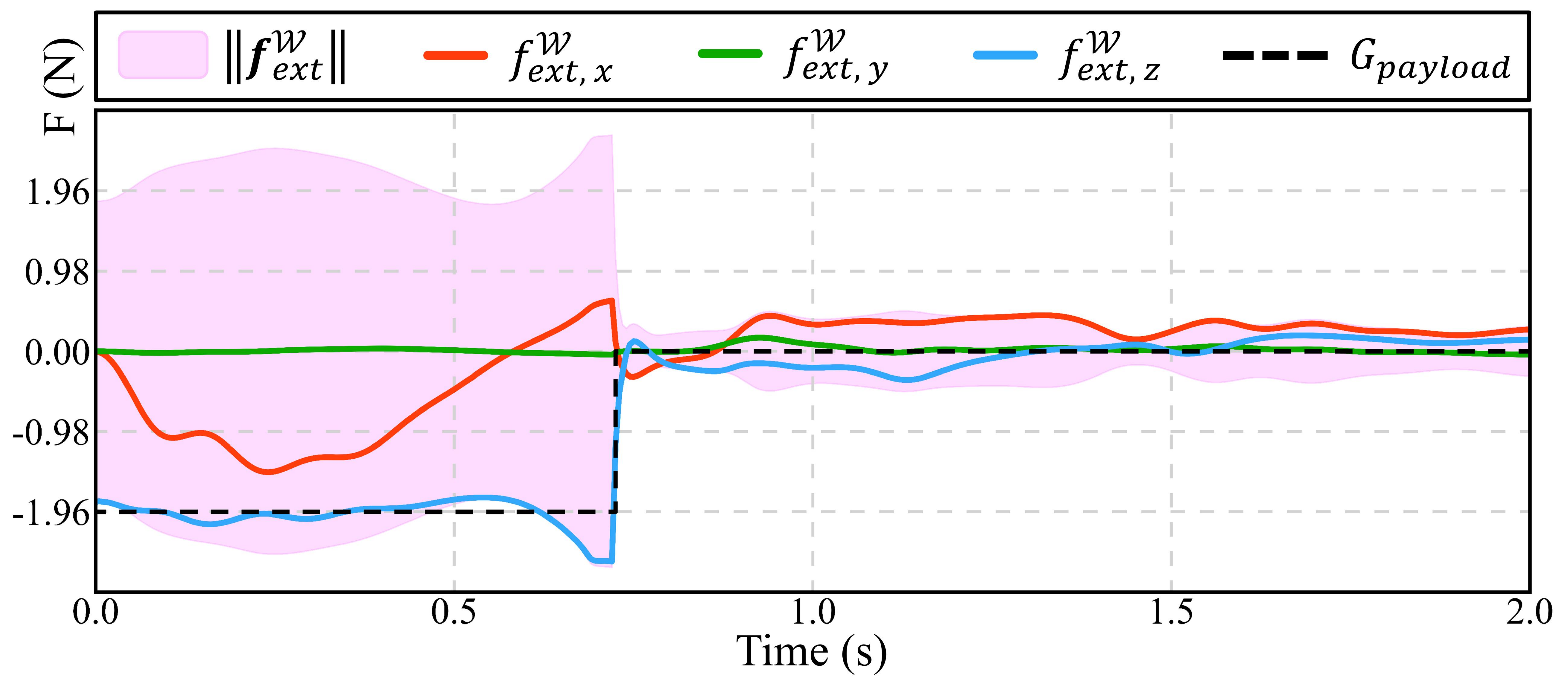}
            \caption{NDOB estimation of linear external disturbance forces during the airdrop process, including the release of a 200~g payload.}
            \label{fig:exp_2}
            \vspace{-5pt}
        \end{figure}

        By recording and analyzing a large amount of real-world experimental data, the results demonstrate that the NMPC control framework, incorporating a hierarchical disturbance observer with NDOB and INDI, leads to a more concentrated airdrop landing point distribution around the reference landing point. Specifically, NDOB compensation of linear forces significantly improves trajectory tracking performance, improving payload landing point concentration by an average of approximately 54\% in scenarios with different release speeds. Furthermore, it is observed that at the moment of release, the NDOB measurement in the horizontal direction changes from -1.96N to 0N, as shown in Fig.~\ref{fig:exp_2}. This enables the proposed autonomous airdrop system to carry payloads of varying unknown masses and effectively handle the model parameter variations caused by releasing payloads with unknown mass.

    \subsection{Comparative Experiment on Throwing Precision}\label{sub:exp3}
\begin{figure}[t]
            \centering
            \includegraphics[width=1.0\linewidth]{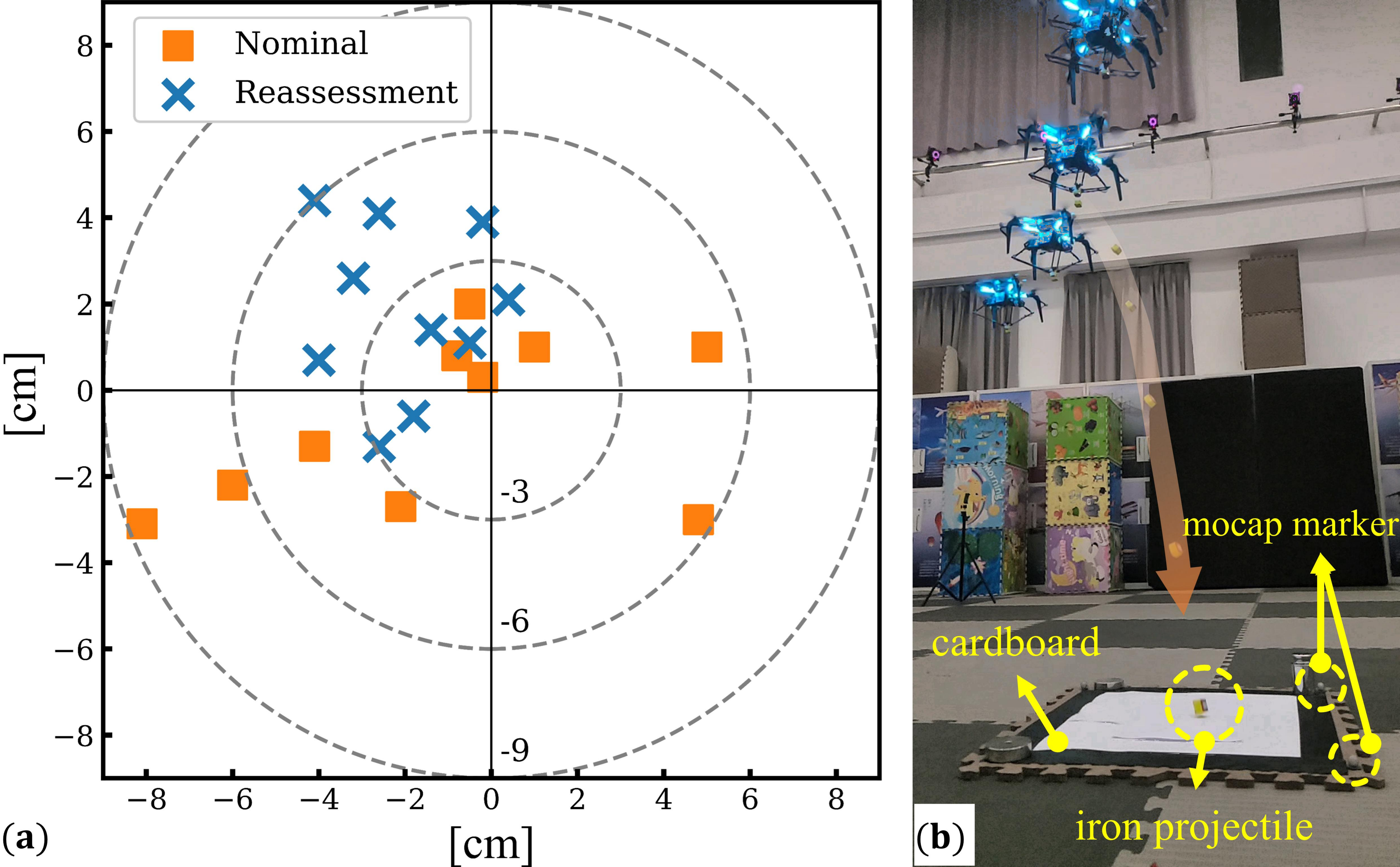}
            \caption{(a): Landing points from ten throws for each of the nominal release trigger method and the proposed prediction-based reassessment approach under the same airdrop trajectory. The reference release velocity was 2.3~m/s. (b): The payload was an iron cube measuring 2.0~cm per side. A cardboard sheet was employed to record the landing positions of the projectile.}
            \label{fig:exp_3}
            \vspace{-10pt}
        \end{figure}

        In this section, we design three distinct flight trajectories and conduct comparative experiments between a nominal release trigger method, which releases the payload strictly at the reference time specified by the planner, and our proposed prediction-based release timing reassessment approach, as illustrated in Fig.~\ref{fig:exp_3}(b). Specifically, the flight trajectories exhibit varying flight states near the payload release point, resulting in different sensitivities to release uncertainty. In the experiment, the AM controller parameters are kept constant, and the hierarchical disturbance compensation framework described above is used for control.  We collect the errors between the actual landing points and the target landing points for both methods and evaluate the performance and feasibility of the proposed approach using MEAN/MAX, as shown in Tab.~\ref{tab:exp_3}.

        \begin{table}[h]
            \centering
            \caption{Landing Precision of Payload In Real Environment}
            \renewcommand{\arraystretch}{1.1}
            \setlength{\tabcolsep}{4pt}
            \begin{tabular}{cc|>{\centering\arraybackslash}p{2.5cm}>{\centering\arraybackslash}p{2.5cm}}
            \toprule
            \multicolumn{2}{c|}{Trajectory}
            & \multicolumn{2}{c}{Landing Precision (MEAN/MAX) [cm]} \\
            \midrule
            \begin{tabular}{c} $v_{r}$ \\ {[m/s]} \end{tabular}
            & \begin{tabular}{c} $a_{\max}$ \\ {[m/s$^2$]} \end{tabular}
            & \text{Nominal} & \text{Proposed} \\
            \midrule
            4.3 & 4.6 & 76.7/83.8 & \textbf{10.1/12.3} \\
            2.3 & 2.7 & 4.2/8.5 & \textbf{3.3/6.0} \\
            2.6 & 2.5 & 14.5/15.3 & \textbf{4.0/4.7} \\
            \bottomrule
            \end{tabular}
            \label{tab:exp_3}
        \end{table}

        Real-world experimental data demonstrate that the proposed reassessment method is effective for various airdrop flights. Although the method ensures landing accuracy even in aggressive airdrop maneuvers, small delays and errors in such aggressive flights lead to significant landing errors. Furthermore, in more moderate airdrop trajectories, the prediction-based approach is able to identify a spatio-temporal state within the NMPC prediction horizon that is more suitable for releasing the payload during actual flight.
        
\section{Conclusion}
    This letter presents an autonomous airdrop system based on an aerial manipulator for precise payload delivery in complex environments. By incorporating trajectory planning with continuous constraints on the target landing point, the system reduces sensitivity to release uncertainty. The control strategy, which includes hierarchical disturbance compensation, enhances robustness and ensures accurate control even with unknown payload mass. Furthermore, utilizing NMPC's predictive capabilities, the system dynamically reassesses the release timing during flight, improving the accuracy of landing. Future work will focus on safe aerial delivery in dynamic environments, encompassing the entire process of payload pickup, transport, and delivery.  

\bibliography{RAL2025_reference} 
\end{document}